# Locally Interpretable Models and Effects based on Supervised Partitioning (LIME-SUP)

Linwei Hu[1], Jie Chen, Vijayan N. Nair, and Agus Sudjianto

Corporate Model Risk, Wells Fargo, USA

June 1, 2018

**Abstract**

Supervised Machine Learning (SML) algorithms such as Gradient Boosting, Random Forest, and Neural Networks have become popular in recent years due to their increased predictive performance over traditional statistical methods. This is especially true with large data sets (millions or more observations and hundreds to thousands of predictors). However, the complexity of the SML models makes them opaque and hard to interpret without additional tools. There has been a lot of interest recently in developing global and local diagnostics for interpreting and explaining SML models. In this paper, we propose locally interpretable models and effects based on supervised partitioning (trees) referred to as LIME-SUP. This is in contrast with the KLIME approach that is based on clustering the predictor space. We describe LIME-SUP based on fitting trees to the fitted response (LIM-SUP-R) as well as the derivatives of the fitted response (LIME-SUP-D). We compare the results with KLIME and describe its advantages using simulation and real data.

## 1  Introduction

Certain classes of supervised machine Learning algorithms have much better predictive performance over traditional statistical methods when used with large data sets. They include ensemble algorithms such as Gradient Boosting Machines (GBMs) and Random Forests (RFs) as well neural networks (NNs). There are also other advantages with large data sets such as automated variable importance analyses and feature engineering. However, these models are very complex and hence hard to interpret without additional tools. In certain environments, such as banking and finance which are regulated, model interpretation is also a critical requirement. For instance, model developers and validators have to understand the model behavior, determine if it is consistent with business knowledge, and so on. Of course, there is also intrinsic interest among modelers in developing insights into the model and explaining the behavior. This has led to a heightened research on global and local diagnostics to interpret complex SML models. In addition, there are also attempts to fit surrogate models that have some loss in predictive performance but are more interpretable. Next, we provide a brief overview of these.

Global diagnostics are aimed at interpreting the overall relationship between input variables and response variable (over the entire model space). These include variable importance analyses, one- and multi-dimensional partial dependence plots (Friedman, 2001) and H-statistics for diagnosing interactions (Friedman & Popescu, 2005), related ones such as ICE (Goldstein, Kapelner, Bleich, & Pitkin, 2013) and ALE plots (Apley, 2016), derivative-based analyses (Kucherenko, 2010; Sobol & Kucherenko, 2009), Sobol indices (Iooss & Lemaitre, 2014), Shapley analyses (Lundberg & Lee, 2017), and so on.

---

[1] email: Linwei.Hu@wellsfargo.com

On the other hand, local diagnostics are aimed at understanding relationship in smaller, local regions, with the idea that a simple parametric model may be used to approximate the input-output relationship. This is the topic of the present paper.

The concept of fitting surrogate models is widely known in the computer experiments literature where it goes by the name of emulators (Bastos & O'Hagan, 2009). In the ML literature, it is referred to as model distillation (Tan, Caruana, Hooker, & Gordo, 2018) and model compression (Bucilua, Caruana, & Niculescu-Mizil, 2006). One of the earliest implementations that we know of is the concept of "born again trees" (Breiman & Shang, 1997). See also the recent work of Vaughan et al. (2018) on Explainable Neural Networks (xNNs). If we piece together the locally interpretable models discussed in this paper to get a global model, they can also be interpreted as surrogate models.

Perhaps the most well-known locally interpretable model currently is LIME proposed by Ribeiro et. al (2016). Given a point in the high-dimensional input space (possibly one of the observations), LIME builds a local model around the point. There are different implementations of LIME in different software packages, and we will not delve into these details of LIME. Rather, our interest is in fitting locally interpretable models on the entire input space. KLIME, a variant of LIME, has been proposed for this purpose by H2O in Hall, Gill, Kurka & Phan (2017). It partitions the input space into K partitions using clustering techniques (typically K-means) and then fits local models within each cluster. The value of K is chosen so that the predictions from all the local models will maximize $R^2$. However, the unsupervised partitioning approaches can be unstable, yielding different partitions with different initial locations. More importantly, the unsupervised partitioning does not incorporate any model information which seems critical to preserving the underlying model structure. Finally, K-means partitions the input space according to the Voronoi diagrams, it is less intuitive in business environment where modelers are more used to rectangle partitioning.

Our approach uses supervised partitioning (tree algorithms) and then fits the local model. We apply this approach to fitting local models to both the fitted response (LIME-SUP-R) and fitted derivatives (LIME-SUP-D). For LIME-SUP-R, we fit parametric model-based trees both to decide on the partitions and to fit the local model within each partition. Thus, if the parametric model is a good fit in a given region, no further partitioning is requited. For LIME-SUP-D, we use the usual piecewise constant trees as it seems most interesting to determine when the partial derivatives change. This can be easily modified to fit model based trees to the derivatives as well. We show that LIME-SUP leads to considerable improvement in performance as well as interpretability over KLIME.

The paper is organized as follows. Section 2 describes the new methods and algorithms in detail. Section 3 analyzes their performance through a simulation study and compares them with KLIME. Section 4 provides comparisons through two real data sets. The advantages of our methods are summarized in Section 5.

## 2 Description of LIME-SUP Methodology

### 2.1 LIME-SUP-R based on Model-Based Trees

The concept of "model-based trees" has been around for more than two decades. It was motivated by the fact piecewise constant fits often produce deep and hard to interpret trees. Several algorithms have been developed using parametric models to split the nodes and also fit the parametric models to terminal nodes. M5 (Quinlan, 1993) is the most prominent representative, followed by algorithms developed by Wei-Yin Loh and his coworkers, for example LOTUS (Chan & Loh, 2004).

We consider here the case where the original dataset for the SML algorithm is partitioned into three: i) training; ii) validation; and iii) testing. The local models are also developed, validated and their performances assessed on these three data sets respectively.

The LIME-SUP-R algorithm works as follows:

1. Let $\{X_{1i}, \ldots, X_{ki}, i = 1, \ldots N\}$ be the set of predictor (independent) variables used to train the original SML algorithm, where $N$ is the number of training dataset observations. These will be used for partitioning and model fitting for the trees.[2]
2. Let $\{\hat{Y}_i, i = 1, \ldots, N\}$ be the fitted responses from the SML algorithm which will be used in the supervised partitioning algorithm. For continuous response, they are the fitted responses, and for binary response, they will be predicted probabilities or logits of the fitted probabilities.
3. For the specified class of parametric model (say linear regression model with no interactions), fit a model-based tree to the responses and predictors in the training dataset. Specifically,
   a) Fit an overall parametric model (from the specified class) at the root node to the (fitted SML) responses and predictors.
   b) Find the best split to partition the root node into two child nodes. This is done by (again) fitting the same class of parametric models to all possible pairs of child nodes and determining the "best" partition. This involves searching over all partitioning variables and possible splits within each variable and optimizing a specified fit criterion such as MSE or deviance.
   c) Continue splitting until a specified stop criterion is met; for example, max depth, minimum number of observations in the child node, or the fit is satisfactory.
4. Prune back the tree using appropriate model fit statistics such as $R^2$, improvement in $R^2$, improvement in SSE, etc. on the validation dataset.

---

[2] In some cases, there may be separate sets of variables $\{X_1, \ldots, X_k\}$ for modeling and $\{Z_1, \ldots, Z_l\}$ for partitioning, and they may be (an important) subset of all variables.

5. Once the final tree is determined, use a regularized regression algorithm (such as LASSO) to fit a sparse version of the parametric model at each node.[3]

There are multiple approaches to Step 3(b) which involves finding the best partitions. One method is to exhaustively search all combinations of partitioning variables and candidate split points (e.g., percentiles), but it is very time consuming. Another approach is the M-Fluctuation test in Zeileis et.al. (2008) which is a fast algorithm to select the best *partitioning variables*, but we found that it may not always pick the variable that minimizes the SSE. Therefore, in our implementation, we use M-Fluctuation test as a filtering step and then apply exhaustive search for the top few variables. This hybrid approach seems to work well.

## 2.2  LIME-SUP-D

The overall approach is the same as LIME-SUP-R but with several key differences:

i) The response(s) for the tree algorithm in Step (2) above are based on the first partial derivatives of the fitted responses (from the SML algorithms):

$\{\frac{\partial \hat{Y}_i}{\partial x_k}, i = 1, \ldots, N;\ k = 1, \ldots, K\}$ and not the fitted responses $\{\hat{Y}_i, i = 1, \ldots, N\}$.

The derivatives have to be scaled appropriately before applying the algorithm. Specifically, before splitting at each node, we compute the standard deviations of each independent variable and multiply the corresponding partial derivatives by these standard deviations. To be precise, let

$$SD_k = std.dev\{X_{k1}, \ldots, X_{KN}\}.$$

Then, we scale the partial derivatives as

$$\{S(\frac{\partial \hat{Y}_i}{\partial x_k}) = \frac{\partial \hat{Y}_i}{\partial x_k} \times SD_k, i = 1, \ldots N,\ k = 1, \ldots, K\},$$

and use these as the $N \times K$ matrix of responses.

ii) There are different ways to fit the multivariate response to the independent variables: i) use multivariate regression techniques to fit the model to the $N \times K$ matrix of responses; or ii) stack the $N$ responses for all $K$ variables into a single vector, repeat the matrix of predictor variables $K$ times and treat it as a single regression problem. The former is more general as it allows us to accommodate the dependence among the columns by specifying a correlation matrix. We used the first approach in our analysis but treated the columns as independent.

iii) The class of parametric models should correspond to one-order lower than those considered for LIME-SUP-R since we are modeling the derivatives. For example, if we consider fitting linear regression models for LIME-SUP-R, then we should fit a piecewise constant tree to each nodes, since the derivatives correspond to coefficients of the liner regression model.

---

[3] We have tried regularized models during the tree-building phase but they tend to be time consuming.

iv) As before, we use the fitted model within each node to get predicted values. In case we are fitting a piecewise constant tree within each node, the fitted value will be the average of all the observations within each node.

Derivatives are readily available for NNs through back propagation. One can use finite differences to approximate the derivatives from the fitted response surfaces for GBM and RF. But these fitted responses tend to be "rough" as they are based on an ensemble of (piecewise constant) trees which are inherently discontinuous. In our framework, we fit a NN surrogate model to the fitted responses of GBM and RF and get the derivatives from the NNs. In separate work to be reported elsewhere, we found that this to work reasonably well. The advantages and limitations will be described in a forthcoming paper.

The tree construction procedures are based on different responses in LIME-SUP-R and LIME-SUP-D, so the two methods will lead to different partitioning. LIME-SUP-R is based on directly optimizing the fit to the response surface, so it generally fits better than LIME-SUP-D, but LIME-SUP-D is much faster if derivatives are available and is more sensitive to change of coefficients. However, despite the difference among the two, they do have a strong connection. Both will continue splitting when the response surface is not well approximated by a linear model and stops splitting otherwise.

## 3 Simulation Study

We use a simple simulated example to assess the performance of LIME-SUP methodologies and compare it with KLIME. We simulated 50,000 binary observations from the following logistic regression model

$$\log\frac{P(Y=1)}{P(Y=0)} = -1 + 0.5x_1 + 1.5(x_2 - 1)_+ - 0.5x_3^2 + 0.5x_4(x_5 + x_6) --Eq\ (1)$$

The independent variables were simulated from independent standard normal. The 50,000 observations were divided into training, validation and testing sets in the following proportions 50%, 20% and 30% respectively. This example has only six independent variables and it is not ideally suited to exploit the predictive power of SML algorithms. Nevertheless, since it is meant to be illustrative, we fitted GBM algorithms to the training data set by tuning the hyperparameters on the validation data. The predicted logits were computed for training, validation, and test data. The first partial derivatives of the logits were obtained by fitting a neural network surrogate model to the GBM results. Figure 10 in the Appendix shows the plots of the derivatives. We can see there is some noise in the derivatives, but the true model is captured quite well.

We next applied LIME-SUP-R, LIME-SUP-D, and KLIME to the responses and the partial derivatives using the steps outlined in the last section. For LIME-SUP-R, we fitted a tree with max depth three leading to eight terminal nodes. For KLIME, we wanted to keep the model fitting steps the same as that for LIME-SUP-R and vary only the partitioning algorithm – unsupervised K-means algorithm vs model-based trees. So we implemented our own version of KLIME as follows:

1. Standardize each of the predictor variables $\{X_1, \ldots, X_K\}$.
2. Keep the number of clusters to be the same as the number of terminal nodes in LIME-SUP (8 here) for fair comparison.

3. Consider several common K-means clustering techniques:

   a. Usual K-means using Euclidean distance $d^2(x_i, x_j) = (x_i - x_j)^T(x_i - x_j)$ (KLIME-E).

   b. Using Mahalanobis distance: $d^2(x_i, x_j) = (x_i - x_j)^T S^{-1}(x_i - x_j)$, where $S$ is the sample variance-covariance matrix (KLIME-M).

   c. K-means after PCA where only the top principal components that account for at least 95% of variation are included (KLIME-P).

Figure 1 shows the tree structures obtained from LIME-SUP-R. The information with each node shows: i) the number of observations, ii) the partitioning variable (with the left child node corresponding to the indicated split and the right child node to its complement), iii) the improvement in SSE after splitting, and $R^2$ value before splitting. The first partition at the root node is on the variable $X_3$, with the split approximately at the value of 0: this tries to account for the quadratic effect. Both partitions at the second level are based on $X_4$, and they are probably accounting for the interaction effect $x_4(x_5 + x_6)$. All four partitions at the next level are based on $X_2$, likely accounting for the non-linear effect (piecewise linear with a single knot at 1). Note that the partitions in node 3 is very similar to node 5 and that for node 4 is very similar to node 6, implying very similar splits of this variable. There is no split on variable $X_1$ as its effect is linear and well captured by the model-based tree. There is no split on variable $X_5$ or $X_6$ either due to depth being shallow here.

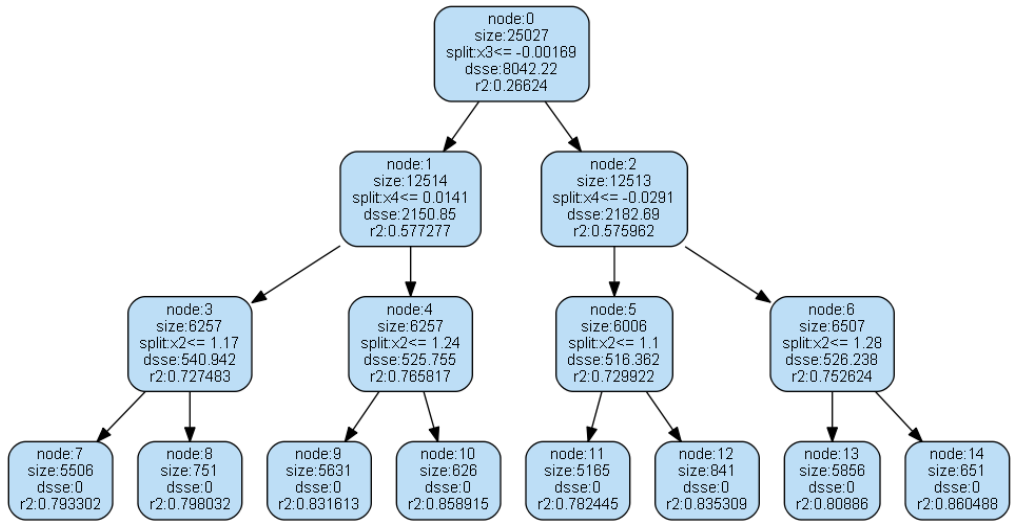

**Figure 1. Tree structure of LIME-SUP-R**

Figure 2 provides a visual comparison of the fitted coefficients for the local linear regression in LIME-SUP-R at the 8 terminal nodes. The effect of $X_1$ is constant across the nodes as it should be, with the values falling roughly around the true value of 0.5 in Equation (1). The values of $X_2$ oscillate roughly between 0 and 1.5: close to 0 for nodes 7, 9, 11, and 13 and close to 1.5 for nodes 8, 10, 12, and 13. This is consistent with the splits on $X_2$ at nodes 3 and 5 vs nodes 4 and 6. The purple line for $X_3$ appears to be an approximately piecewise linear approximation of the quadratic function in Equation (1). Note that this split occurs at the top level, so nodes 7, 8, 9, and 10 share the same split of $X_3$ while remaining nodes correspond to the complementary split. Variable $X_4$ has approximately the same values although it interacts with $X_5$ and $X_6$. Variable $X_4$ was selected over $X_5$ and $X_6$ as it interacts with both. The regression coefficients of $X_5$ and $X_6$ vary between low (nodes 7 and 8; nodes 11 and 12) to high (nodes 9 and 10 and nodes 13 and 14), depending on value of $X_4$. Tree partitions can capture interaction effects and nonlinear effects even if the local model does not include interactions or nonlinearity, but they will not be recovered exactly.

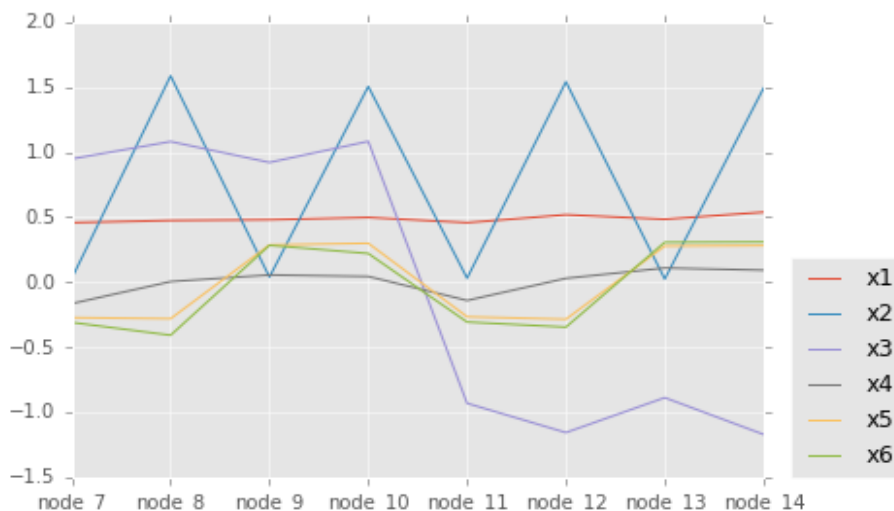

**Figure 2. Coefficients for terminal nodes in LIME-SUP-R**

Figure 3 portrays the regression tree obtained by LIME-SUP-D based on constant local regression models within each node. In this simple example, the partitioning variables and final partitions are almost the same as LIME-SUP-R but this is not typically the case.

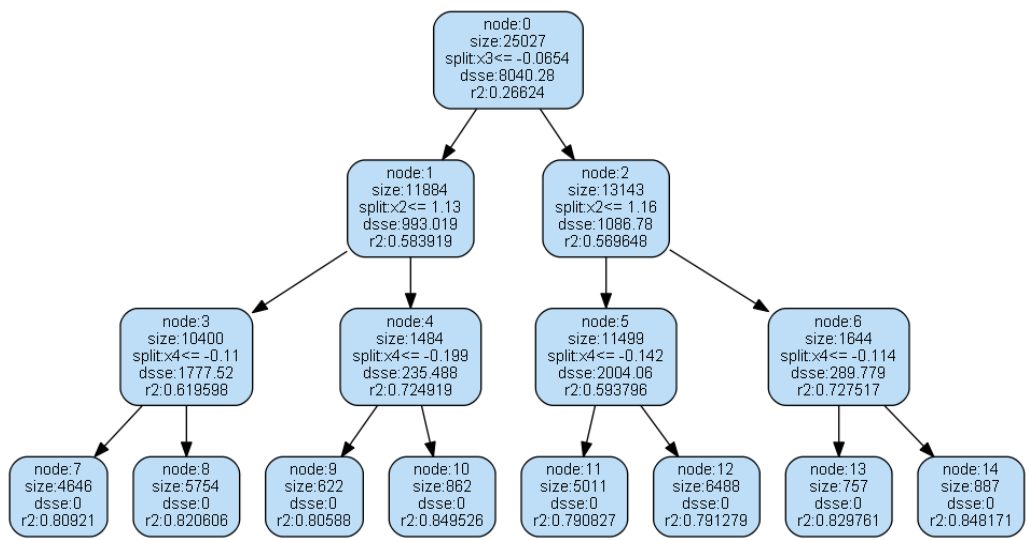

**Figure 3. Tree structure of LIME-SUP-D**

The corresponding coefficients (derivatives) at the terminal nodes are shown in Figure 4**Error! Reference source not found.**. The conclusions are very similar to those in Figure 2.

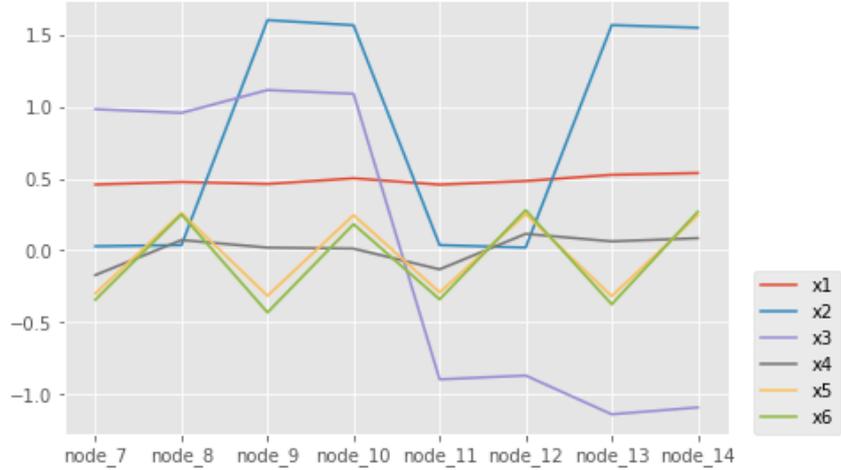

**Figure 4. Coefficients for terminal nodes in LIME-SUP-D**

Figure 5 provides a plot of the regression coefficients for the different clusters from KLIME-E. The values for $X_1$ are more or less the same across the clusters. This is consistent with the model in Eq(1). But there is no clear interpretation of the regression coefficients for the other variables and the model in Eq(1). In fact, the input variable space is a multivariate Gaussian distribution instead of a mix of clusters, so it is not meaningful to apply clustering.

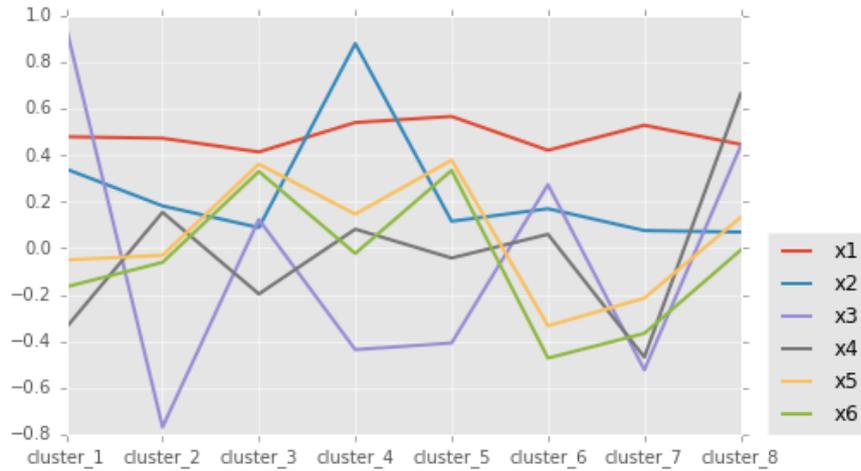

**Figure 5. Coefficients for clusters in KLIME-E**

Table 1 provides a different comparison of the various techniques based on MSE, $R^2$ and AUC of the overall fitted model. They are computed from the logits and KLIME/LIME-SUP fits. We see that two LIME-SUP approaches perform better than all three versions of KLIME in terms of all metrics. With a shallow tree of only three layers, $R^2$ has improved from 0.27 to 0.83.

**Table 1. Mode fit summaries on testing data**

|  | LIME-SUP-R | LIME-SUP-D | KLIME-E | KLIME-M | KLIME-P |
|---|---|---|---|---|---|
| MSE | 0.172 | 0.172 | 0.326 | 0.339 | 0.385 |
| $R^2$ | 0.831 | 0.830 | 0.679 | 0.666 | 0.621 |
| AUC | 0.734 | 0.732 | 0.715 | 0.711 | 0.702 |

Figure 6 provides a different view of the comparisons: values of MSE and $R^2$ computed within each of eight local regions. On the left panels in the top and bottom rows, the 8 partitions correspond to those obtained from KLIME-E (which had the best performance among the three KLIME methods – see Table 1). The partitions on the right panels correspond to those from LIME-SUP-R. Thus, the comparisons of MSE and $R^2$ among the various methods were made among the <u>same</u> partitions: the left panels correspond to the best partitions for the KLIME-E method and the right panels to the LIME-SUP-R method. The conclusions from the right panels may not be surprising: after all, the partitions were best for LIME-SUP-R. But those from the left panels are rather startling: the LIME-SUP methods outperform the KLIME methods for all partitions even though they correspond to the best clusters for KLIME-E! The performance of LIME-SUP-R and LIME-SUP-D are comparable here. In general, we found LIME-SUP-R to yield better interpretation and fits than LIME-SUP-D.

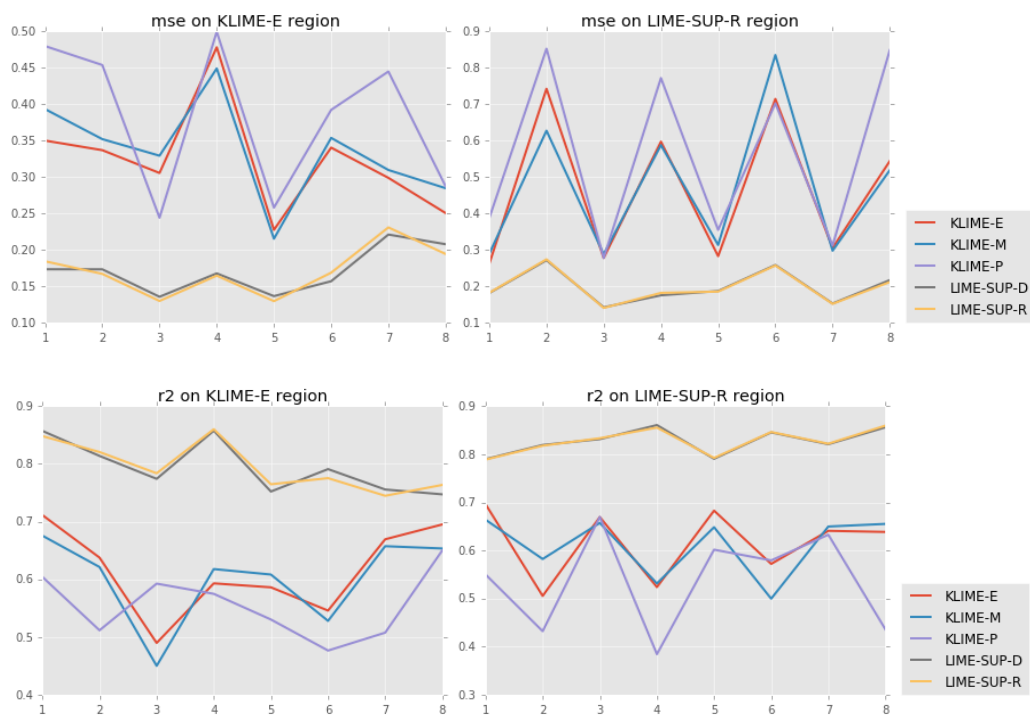

Figure 6. Comparisons decomposed by partitions: Top panel: MSE; Bottom panel $R^2$

## 4 Comparisons on Two Real Applications

In this section, we show selected results from applying the methods to two real examples. The details are proprietary, so we provide only global comparisons.

### 4.1 Personal Line and Loan Data

In this dataset, there were about 37,000 observations and more than 500 variables. The response variable is a binary indicator variable (default or not). The data was divided into training, validation and testing data sets, with about 50%, 20% and 30% of the observations. We first fitted a GBM model with all the 500+ variables on the training dataset and tuning the hyperparameters on the validation dataset. Derivatives were obtained by fitting a neural network to the fitted GBM response surface (details are omitted).

The top 20 variables were then selected using variable importance scores. We included 4 additional variables based on subject matter expertise for a total of 24 variables for LIME-SUP and K-LIME. For LIME-SUP, we fitted a tree with max depth four and pruned it back using minimal reduction in validation SSE. The number of clusters in KLIME is fixed to be the same as in LIME-SUP. We fitted all final local linear models (both LIME-SUP and KLIME methods) using LASSO to get sparse models.

Table 2 provides a comparison of the overall fit metrics: MSE, $R^2$ and AUC. We see that the LIME-SUP methods outperform the KLIME methods. Figure 7 compared the MSE and $R^2$ across partitions, in the same way as we did in Figure 6. Again, we reach a similar conclusion.

Table 2. Comparison of the various methods on testing data

|       | LIME-SUP-R | LIME-SUP-D | KLIME-E | KLIME-M | KLIME-P |
|-------|------------|------------|---------|---------|---------|
| MSE   | 0.143      | 0.159      | 0.239   | 0.256   | 0.240   |
| $R^2$ | 0.754      | 0.727      | 0.588   | 0.559   | 0.587   |
| AUC   | 0.740      | 0.737      | 0.719   | 0.723   | 0.719   |

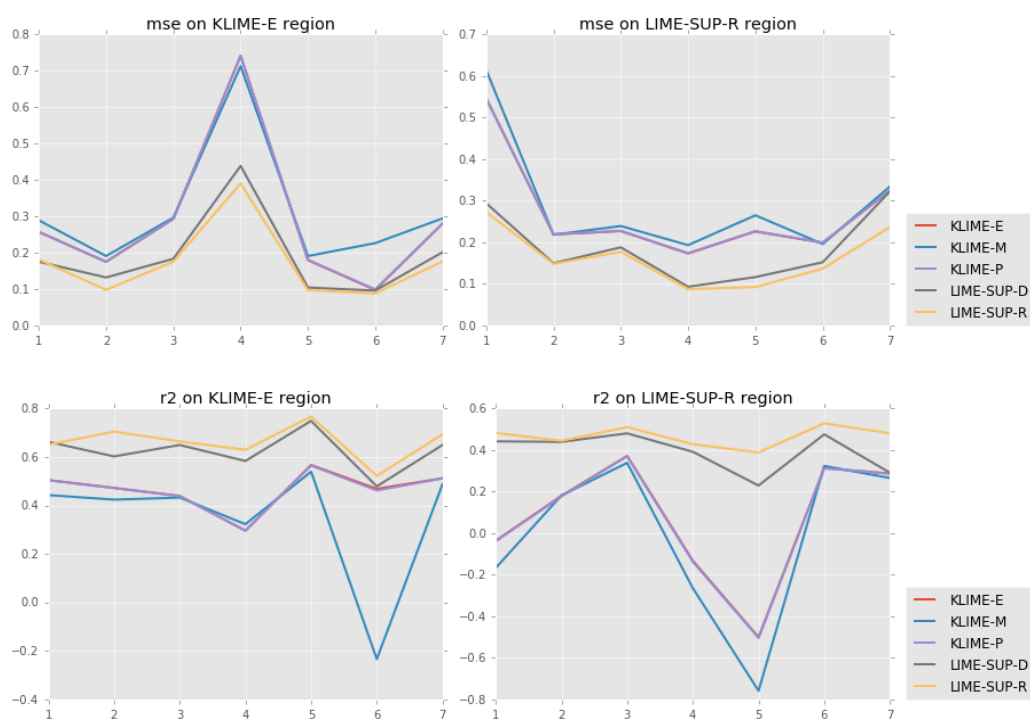

Figure 7. Comparisons for PLL data by partitions: Top panel: MSE; Bottom panel $R^2$

## 4.2 Home Lending Data

This dataset is based on home lending for residential mortgage. For illustrative purposes, we used a randomly selected subset of one million from the original 200M observations. For simplicity, we restricted attention to same subset of 7 predictor variables that were used by the lines of business. Again, the original SML algorithm was GBM. We only provide the essential details here and skip the others; the rest are very similar to the descriptions in the analysis of the PLL dataset. The number of levels for the trees was fixed at three for a total of eight terminal nodes.

Figure 8 is a plot of the fitted coefficients from LIME-SUP-R. We see that the main differences across the terminal nodes are in variables 2 (LTV_Forecast) and variable 3 (delinquency status)[4]. The $R^2$ values in the terminal nodes varied from 0.88 to 0.97 with most of the values in the 0.9+ range. So the underlying local models provide a good fit and the interpretations are quite simple. The results from LIME-SUP-D were a little different. We are pursuing further research to understand these differences.

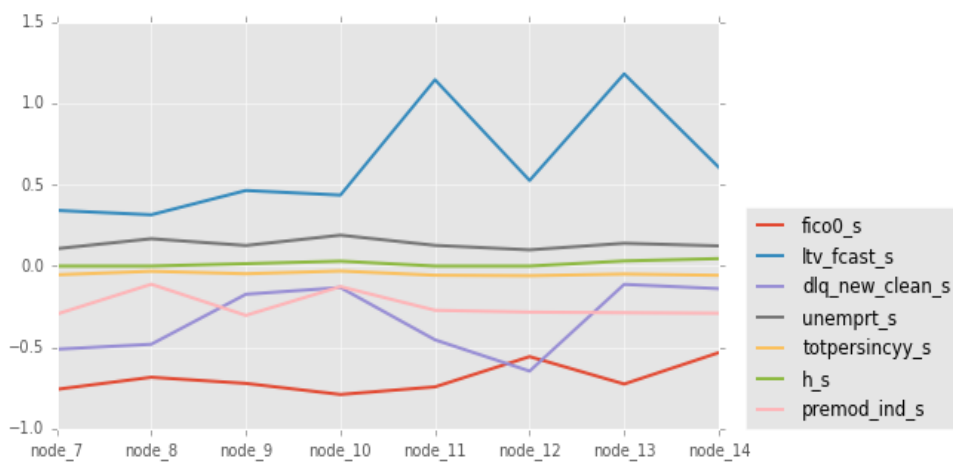

Figure 8. Coefficients at the terminal nodes in LIME-SUP-R for Home Lending Data

Table 3. Mode fit on testing data for Home Lending

|  | LIME-SUP-R | LIME-SUP-D | KLIME-E | KLIME-M | KLIME-P |
|---|---|---|---|---|---|
| MSE | 0.0429 | 0.0552 | 0.0681 | 0.0702 | 0.0647 |
| $R^2$ | 0.975 | 0.967 | 0.960 | 0.959 | 0.962 |
| AUC | 0.834 | 0.833 | 0.831 | 0.832 | 0.832 |

Table 3 provides a comparison of the of the overall MSE and $R^2$ values and they lead to a similar conclusion of the relative performances of the LIME-SUP methods vs KLIME methods, although the differences here are not striking as before. Figure 9 shows the comparisons within individual partitions. It is similar to Figure 7 except the left panels are based on the best clusters for KLIME-P which had the best performance within that group. The conclusions are similar to those in Figure 7, except for the fact that LIME-SUP-D has worse performance than KLIME methods for some nodes. Again, LIME-SUP-R outperforms KLIME methods on even the left panels where the clusters are based on KLIME-P.

---

[4] The suffix "_s" in the variable name indicates this variable is standardized to have variance 1.

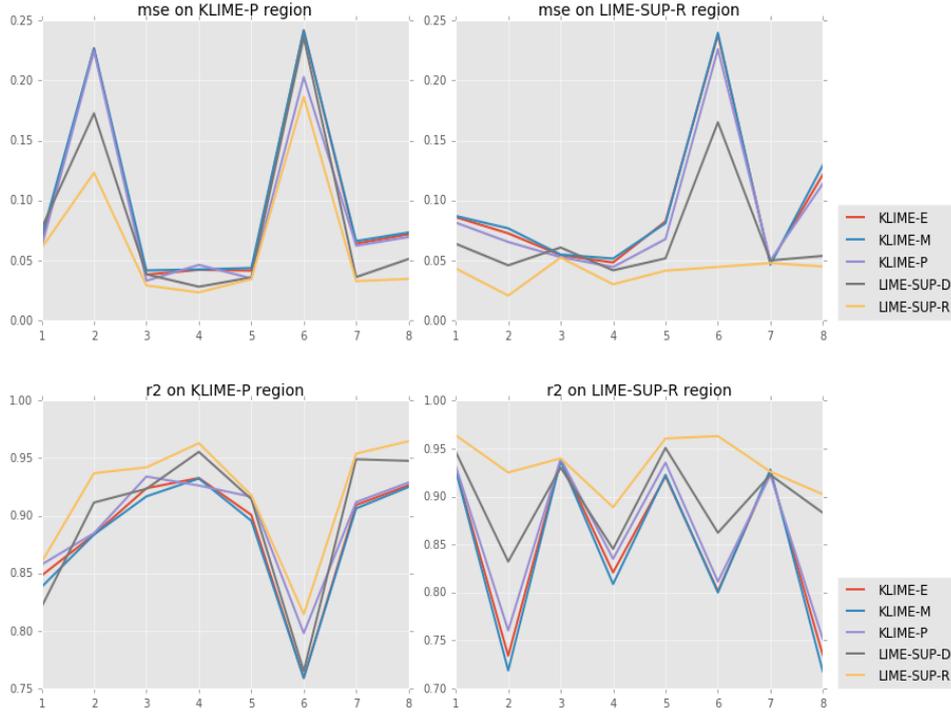

Figure 9. Comparisons for Home Lending data by partitions: Top panel: MSE; Bottom panel $R^2$

## 5 Summary

We have introduced two classes of locally interpretable models and effects based on supervised partitioning: i) LIME-SUP-R and ii) LIME-SUP-D. Our investigations show that both of them perform better than KLIME methods. Further, LIME-SUP-R appears to have slightly better performance in terms of predictive accuracy and interpretability. We are currently undertaking further work to understand the differences and develop insights.

In conclusion, LIME-SUP has the following advantages over KLIME methods.

- The supervised partitioning leads to the use of the underlying model structure in developing the partitions. Thus, it approximates the original model better than KLIME and typically leads to more meaningful partitions. If the underlying local model is linear, LIME-SUP will not split the node further so it is more economical. It attempts to capture any nonlinearity or interactions through partitioning, and the splits are optimized to tell accurately where the nonlinearity or interaction happens, so it is more interpretable. Of course, one can fit higher-order local models that can incorporate quadratic nonlinear effects and simple interactions directly.
- In our experience, supervised partitioning leads to more stable trees. In addition, the tree structure is easy to understand; its hierarchical structure lays out the most important, second most important segmentation feature, and so on.
- The upper level tree nodes offer a semi-global level interpretation. They provide a model segmentation scheme with a small number of segments.
- In addition to being an interpretation tool, LIME-SUP can also be viewed as a technique for constructing global model-based trees. Since we are building the trees on fitted responses (from GBM, RF and NN) rather than the original response variable, there is less noise, so there are fewer concerns on stability of the trees or overfitting, even for deeper trees.

## A. Appendix

Figure 10 shows the plots the derivatives. We can see there is some noise in the derivatives, but the true model is captured quite well, since

- The derivatives of $x_1$ are around the true constant value 0.5.
- The derivatives of $x_2$ are around 0 before 1 and increases after 1.
- The derivatives of $x_3$ depend linearly on $x_3$, consistent with the quadratic effect.
- The derivatives of $x_4$ depend linearly on $x_5 + x_6$.
- The derivatives of $x_5$ and $x_6$ depends linearly on $x_4$.

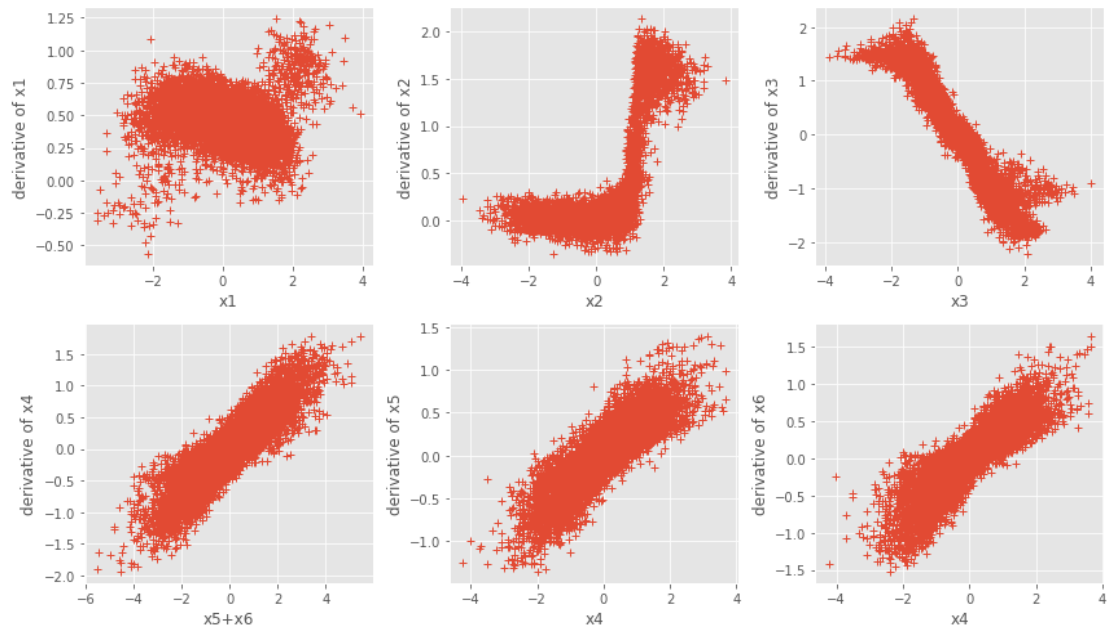

Figure 10. Derivatives of input variables for the simulation study